# Deep-and-Wide Learning: Enhancing Data-Driven Inference via Synergistic Learning of Inter- and Intra-Data Representations [*]

Md Tauhidul Islam[1] and Lei Xing[1,2,3,*]

[1]Department of Radiation Oncology, Stanford University, Stanford, California-94305, USA
[2]Department of Electrical Engineering, Stanford University, Stanford, California-94305, USA
[3]Institute for Computational and Mathematical Engineering, Stanford University, Stanford, California-94305, USA

## ABSTRACT

Advancements in deep learning are revolutionizing science and engineering. The immense success of deep learning is largely due to its ability to extract essential high-dimensional (HD) features from input data and make inference decisions based on this information. However, current deep neural network (DNN) models face several challenges, such as the requirements of extensive amounts of data and computational resources. Here, we introduce a new learning scheme, referred to as deep-and-wide learning (DWL), to systematically capture features not only within individual input data (intra-data features) but also across the data (inter-data features). Furthermore, we propose a dual-interactive-channel network (D-Net) to realize the DWL, which leverages our Bayesian formulation of low-dimensional (LD) inter-data feature extraction and its synergistic interaction with the conventional HD representation of the dataset, for substantially enhanced computational efficiency and inference. The proposed technique has been applied to data across various disciplines for both classification and regression tasks. Our results demonstrate that DWL surpasses state-of-the-art DNNs in accuracy by a substantial margin with limited training data and improves the computational efficiency by order(s) of magnitude. The proposed DWL strategy dramatically alters the data-driven learning techniques, including emerging large foundation models, and sheds significant insights into the evolving field of AI.

The past decade has witnessed remarkable advancements in deep learning and artificial intelligence (AI)[1–6]. While deep learning's ability to extract knowledge from data has driven breakthroughs across numerous fields, significant challenges in current DNN models still need to be addressed[7–12]. Generally, training a DNN is to find the optimal nonlinear function that maps an input manifold, extended by the input data, to an output manifold, defined either by the labels for supervised learning or by other criteria for unsupervised learning[13,14]. While a holistic view of manifold mapping across the entire input dataset is highly desirable at each stage of learning, feature extraction in current deep learning typically proceeds in a batch-by-batch fashion without explicitly considering the relationships among the input data as a whole. This approach significantly compromises the data-driven learning process—not only greatly slowing down training but also leading to suboptimal solutions.

In this work, we propose a mechanistic strategy for learning inter-data relationships and leveraging these interactions for enhanced data-driven modeling while improving the computational efficiency. The proposed learning scheme, called deep-and-wide learning (DWL), captures essential features both within individual data points (intra-data features) and across data points (inter-data features). In our proposed framework, the latter are characterized by low-dimensional (LD) components of the dataset and provide critical insights into the overall structure of the input data. To implement this novel DWL scheme, we introduce a dual-interactive-channel network (D-Net). In contrast to conventional DNNs, which focus on HD feature extraction from individual input data using multiple layers, our D-Net extracts both high- and low-dimensional feature representations from the dataset and utilizes them synergistically for substantially improved learning. The LD representation, derived by using a Bayesian dimensionality reduction (BDR) approach, reveals the overall relationships among the data and helps to identify more suitable HD representations at different layers. Computationally, the BDR approach first learns the distributions of the input data samples in a HD manifold space. It then leverages the acquired data distribution information to produce an LD representation that preserves structural integrity between input manifold space and LD space (Fig. 1 a). The HD representation in D-Net is generated from conventional feature extraction operations, which capture the contextual details of each input data. These two representations are integrated through a dense layer, where the HD features are combined with the LD information to ensure consistency across all feature levels. In this way, the feature information learned by the convolutional layers is refined by the LD components, and vice versa. This interactive process improves the network's performance

---

while mitigating the effects of noise and data biases. With the integration of LD features, superior learning performance with significantly faster training convergence become possible, even with only a single convolutional block (Fig. 1 b).

We tested the D-Net on a broad array of publicly available datasets including image, text, and genomic data—across various domains. Our results demonstrate that D-Net not only significantly outperforms state-of-the-art DNNs in inference accuracy but also in computational efficiency. The proposed DWL has the potential to transform the development of data-driven models in the rapidly advancing field of AI.

## Results

### DWL speeds up the fine-tuning process of large language models (LLMs) by approximately 200-fold, while nearly doubling their inference accuracy

LLMs have recently garnered significant attention. Due to their vast number of parameters, training and fine-tuning these models are computationally intensive. The effectiveness of our D-Net approach is initially demonstrated using two open-source foundation models: BERT[15] and GPT2[16]. We select a standard text dataset consisting of factory reports from MATLAB (Mathworks Inc., Natick, MA, USA). The dataset contains a table containing 484 reports with various attributes, including plain text descriptions and categorical labels. We obtained pretrained BERT and GPT2 models and use them with a standard classifier[15] to categorize the factory reports. The pretrained models were frozen, and only the classifiers were fine-tuned during our experiments. The data was split into training, validation, and testing sets in a 70%:15%:15% ratio.

We visualize the feature distributions of the final fully connected layer (FCL) of the BERT and Dual-channel BERT (D-BERT) classifiers in the first row of Fig. 2, using t-SNE[17] in columns 1 and 2, respectively. The features generated by BERT are not well-separated among the different classes. In contrast, D-BERT produces features that become visually well-separated and exhibit a Gaussian distribution (column 2, Fig. 2). This improvement in feature separation is further validated by the quantitative metric in Fig. S7, which shows a significant increase in the adjusted Rand index (ARI)[18] after each intermediate layer of the networks. The performance comparison of the two networks is presented in rows 3 and 4 for epochs 10 and 2,000, respectively. Remarkably, D-BERT achieves nearly 88% accuracy in text classification after just ten epochs, while BERT reaches only 42% accuracy at the same stage. Moreover, BERT plateaus at 69% accuracy after 2,000 epochs, whereas D-BERT improved slightly to 90%. It is intriguing that D-BERT's calculation efficiency is enhanced by approximately 200-fold compared to BERT, due to the inclusion of both inter- and intra-data interactions (Fig. 3). Note that the computational time of BDR in DWL is negligible (see supplementary section 9).

The D-Net framework demonstrates even more striking performance with GPT2. The second row of Fig. 2 presents the feature distributions of the FFC layer for GPT2 and Dual-channel GPT2 (D-GPT2) classifiers, using t-SNE in columns 1 and 2, respectively. GPT2's features appear almost entirely randomly distributed, while D-GPT2 produces clearly separated features (column 2, Fig. 2). In terms of performance, GPT2 achieves around 41% and 43% accuracy after 10 and 2,000 epochs, respectively. Meanwhile, D-GPT2 reaches 85% accuracy at 10 epochs and 92% at 100 epochs. Once again, the fine-tuning efficiency of D-GPT2 is approximately 200 times greater than that of GPT2.

### DWL achieves highly accurate classification of tumors in MRI images

The efficacy of our approach is demonstrated by comparing the performance of D-Net with the existing VGG-19 network[19]. For training of the models, we selected an MRI dataset consisting of 3,064 T1-weighted contrast-enhanced MRI images collected from 233 patients (1,025 sagittal images, 994 axial images, and 1,045 coronal images)[20]. This dataset includes three types of tumors: meningioma (708 images), glioma (1,426 images), and pituitary tumor (930 images). We present the feature distribution of the VGG-19 in the first column of Fig. 4 using t-SNE (row-1). It can be seen that the features are not well separated among different classes. In contrast, for our D-Net with only one block of convolutions, the features become well separated and exhibit a Gaussian distribution as seen in the second column of Fig. 4. In addition to the visual improvements (Fig. 4), the separation among data classes is reflected in the quantitative metrics shown in Fig. S7, where the data distribution quality, as measured by the ARI index, improves after each intermediate layer within the networks. The results suggest that D-Net enhances the extraction of discriminative information from the data while reducing information loss and noise.

The performance of the two networks is shown in the second and third rows of Fig. 4 for epochs 5 and 100, respectively. Remarkably, D-Net achieves nearly 77% accuracy in tumor classification after just five epochs, while VGG-19 only reaches 35% accuracy at the same stage. Even after 100 epochs, VGG-19 plateaus at 63% accuracy, whereas D-Net maintains its 77% performance. It is worth noting that as the BDR algorithm in D-Net has no learnable parameters, the network size is thus effectively reduced by 10-fold. Consequently, the efficiency of training computation is enhanced by a factor of approximately 20.

**DWL enables highly accurate annotation of cell types in single-cell RNA-seq (scRNA-seq) data analysis**

The potential of our D-Net is also illustrated by using a Tabula Muris (TM) scRNA-seq dataset (transcriptomics of 20 mouse organs, containing 55 different cell classes)[21]. The gene expression data of 54,865 cells, each with 19,791 genes, are analyzed. A state-of-the-art DNN architecture named ActiNN[22] was trained on the TM dataset for cell type annotation. The t-SNE visualization of the feature distributions at the last three FCLs is shown in columns 1-3 in the first row of Fig. 5. It is seen that most data classes are not well-separated in all three FCLs. In contrast, the t-SNE visualization of the features in the last three layers of D-Net shows completely separated data classes. The classification performance of the two networks is shown in the third and fourth rows for epochs 1 and 100, respectively. Computationally, D-Net reduces the training time by a factor of 100. It is worth of noting that the proposed DWL approach dramatically reduces the need for large size of training data. For 50%, 70%, and 90% of the training data, ActiNN achieves accuracies of 86%, 85%, and 87%, respectively, after 100 epochs of training. Remarkably, D-Net achieves 92%, 94%, and 97%, respectively, with only a single epoch (after one epoch, ActiNN achieves accuracies of only 15%, 43%, and 51%, respectively).

**Incorporation of LD features dramatically boosts the performance of existing deep learning models**

The principle of D-Net architecture design is quite broad and applicable to greatly augment the performance of any existing DNNs. To illustrate this, we first use a mammography dataset from the Breast Cancer Digital Repository (BCDR) to study several representative DNNs. The BCDR dataset includes mammography images from 1,734 patients, all annotated by radiologists.

We present the feature distribution of a trained AlexNet network using the BCDR dataset in the first column of Fig. 6, where the features are not well-separated among different classes. However, after integrating the LD feature branch into AlexNet (Fig. 6), the separations among features from different classes are significantly increased, as shown in the second column of Fig. 6. The classification performance of traditional DNNs and their D-Net counterparts is shown in the second row of Fig. 6. The accuracy of AlexNet, VGG16, EfficientNet, and ResNet is 67%, 78%, 63%, and 91%, respectively. Remarkably, the corresponding D-Net models achieve accuracies of 89%, 90%, 76%, and 99%, respectively. Furthermore, the training times for all these DNNs are reduced by more than 5 times. This substantial improvement highlights the significant role of LD feature components in enhancing the learning performance.

We further demonstrate the impact of LD features using the ISIC-2019 dataset, which includes 25,331 dermoscopic images across various categories: 4,522 melanoma, 12,875 melanocytic nevus, 3,323 basal cell carcinoma, 867 actinic keratosis, 2,624 benign keratosis, 239 dermatofibroma, 253 vascular lesions, and 628 squamous cell carcinoma cases[23,24]. Three representative DNNs and their respective D-Net counterparts were trained on this dataset for skin cancer diagnosis. The t-SNE visualizations of the feature distributions extracted from the training and testing data at the final FCL of existing DNNs are shown in columns 1 and 2 of the first row of Fig. 7. Again, the data classes are not visually distinguishable, and the same is reflected in the classification accuracy bar plots in the third row. In contrast, after integrating the LD channel with these networks, the t-SNE visualizations show fully separated data classes. The classification performance of the two types of DNNs is also shown in the 3rd row: AlexNet, Xception, and GoogleNet achieve accuracies of 63%, 72%, and 78%, respectively, whereas their dual-channel counterparts reach accuracies of 73%, 85%, and 87%, respectively.

Supplementary sections 3 and 4 further illustrate the superior performance of D-Net in five additional classification and regression tasks across various datasets, including TCGA[25,26], COVID-19[27], colonoscopy[28], mammography[29] and scRNA-seq datasets. Across all these tasks, D-Net consistently stands out in performance when compared to the existing DNNs.

## Discussion

With the ever-increasing applications of AI across diverse fields such as healthcare, language processing, autonomous driving, finance, and smart homes, the demand for rapid processing and reliable learning algorithms has become paramount. While current deep learning models are undeniably powerful, they are also computationally intensive, susceptible to noise, and often difficult to interpret, making it challenging to fully understand how these models generate their predictions. In practice, despite its algorithmic variety and complexity, data-driven inference ultimately involves learning essential features from the input data and reliably discerning patterns from these extracted features. Currently, this is achieved by nonlinearly transforming the input data through the extraction of HD features, followed by dimensionality reduction and reparameterization. In this work, we highlight the significant role of inter-data relationships, characterized by the LD feature representation of the input data, and introduce the concept of DWL to leverage the LD features and their interactions with the HD feature space for substantially improved learning and inference. Along these lines, we have developed a D-Net architecture for DWL by effectively combining the LD and HD features during learning. We demonstrate that, across a wide variety of applications, the proposed D-Net architecture significantly accelerates computation, improves accuracy, and enhances the learning process. Generally, LD features are more interpretable as they often capture the fundamental structure and relationships within a dataset. By integrating these LD features into the learning process, we are able to regularize the extraction and utilization of HD features, enhancing the learning mechanism. As demonstrated in the Results section, this integration not only improves model accuracy but also significantly increases computational efficiency. Furthermore, monitoring the evolution of LD components and their interactions with HD features throughout the learning process offers valuable insights into how the model adapts and learns, thus enhancing the overall interpretability of the data-driven framework (see Figs. S6 and S8). This dual-feature approach not only refines the model's decision-making process but also provides a better understanding of the underlying feature data dynamics, making the learning models more transparent and reliable.

Humans are generally more adept at understanding LD features in a given dataset[30]. It is well-established that humans possess a remarkable ability to learn and recognize objects based on limited information, identifying key similarities and differences after only a few examples[31]. This cognitive efficiency allows people to retain essential features while distinguishing important nuances between objects. In contrast, current neural networks typically rely on a brute-force extraction and processing of HD features, exploring countless combinations of features in order to learn these relationships. By incorporating meaningful LD feature representation that encapsulates the critical relationships of the data, DWL greatly enhances the learning process. Indeed, by guiding the model with the features distilled by the LD representation of the input data, DWL shows remarkable performance in learning from data. This approach allows the network to focus on the most relevant information, much like the human learning process, thereby enhancing the model's learning and generalization process.

DWL integrates both LD and HD data features and leverages their interaction in the learning and inference processes, allowing it to discover meaningful patterns and insights from smaller datasets without sacrificing accuracy. Indeed, as shown in our analyses, DWL excels in performance with much reduced need for training data as compared to conventional models, making it a valuable solution in many data-constrained environments (see Fig. 5). By reducing the need for extensive data collection, annotation, and preprocessing—often time-consuming and costly—this approach offers transformative potential across various domains, particularly in healthcare, scientific research, and industries where data is scarce or difficult to obtain. Additionally, DWL's ability to generalize efficiently from limited data helps mitigate issues like overfitting and model bias, which are common in conventional models applied to smaller datasets. Thus, DWL paves the way for faster deployment of AI technologies, enhancing their accessibility and impact.

To gain further insights into the role of LD features in facilitating the learning process, we examined the influence of the number of BDR components on DWL performance. The results for the text and tumor classification tasks are presented in Supplementary Section 7. Interestingly, even with a low number of components (e.g., 2), DWL substantially improves the performance of traditional DNNs, though the reduction in computational time required to train the network is not as significant as when a larger number of BDR components are used. In this case, the HD features are largely mixed across classes (Fig. S9). However, as more LD components are introduced, their influence on HD feature extraction increases dramatically, leading to improved performance and computational speed (Figs. 8, S11 and S10). Additionally, as shown in Supplementary Section 7, the training process becomes more stable as the number of BDR components increases, with less variation in training and validation accuracy across epochs. We observed that this improvement plateaus after a certain point (e.g., 200 components). Therefore, using a number of BDR components at or slightly beyond this threshold is generally desirable for achieving high performance and faster computation. In our implementation, a Bayesian approach with automated determination of the necessary number of components is incorporated (see Methods).

In our D-Net implementation, we have a Bayesian approach for extracting LD components of the data. It is worth noting that this LD feature extraction can also be achieved using other classical machine learning methods, such as PCA[32,33], MDS[34], LDA[35], and ICA[36]. Extensive ablation studies have been conducted, and the results indicate that, while these classical approaches do enhance learning, their performances are not as good as BDR (see section 8 of the supplementary materials). In the current D-Net architecture, interactions between LD and HD features are integrated together through a FCL. Other ways to model these interactions, such as incorporating higher-order relationships between LD and HD features, may further improve the performance of DWL, leading to even more accurate and robust models.

Extensive experiments on various benchmark datasets were conducted to validate the efficacy of our D-Net approach. The results show that the integration of LD features improves information flow within the network and enhances its resilience to noise in input data—an important advantage for real-world applications. D-Net also demonstrates superior generalization to unseen data by effectively capturing the underlying structure of the dataset. In our implementation, the DWL operates in a dual-channel fashion. However, the network design can also be realized by using a single-channel framework, where LD representations are computed after HD features are extracted at a number of selected layers. This ensures that HD and LD features remain consistent throughout the network during training. Results from one such implementation are shown in Fig. S13 for the scRNA-seq dataset, demonstrating that both implementations yield similar results. Finally, we note that D-Net should be extendable for integration into other learning paradigms, such as generative, reinforcement and unsupervised learning, to enhance their performance.

## Outlook

As AI becomes increasingly integrated into daily life, the importance of computational speed and reliability grows to meet the expanding demands of modern society. The proposed DWL framework, realized through the D-Net architecture or similar models, marks a significant advancement in data-driven modeling. By combining HD and LD feature representations, DWL not only enhances the accuracy, computational efficiency, and interpretability of learning models, but also improves their robustness and generalization ability across a broad range of applications. Faster computations, paired with reduced demands for extensive training data, enable DWL systems to analyze data quickly, while conserving energy. This novel approach holds great potential for addressing complex challenges in AI, delivering superior performance across various disciplines.

## Methods

## Theory

This section outlines the mathematical operations of the DWL that processes two distinct types of features (HD and LD), $F_h \in \mathbb{R}^n$ and $F_l \in \mathbb{R}^m$, through separate channels. The extracted features are then aggregated and processed for tasks such as classification or regression.

### 1. Feature Extraction
*Channel 1: HD feature extraction*

For input $X$, the features extracted by the first channel can be written as

$$F_h = f_h(X; \theta_h)$$

where $f_h$ is the function representing the operations in the first channel (e.g., a series of convolutions and activations), $\theta_h$ represents the parameters of these operations, and $F_h$ is the extracted feature set from $X$.

*Channel 2: LD feature extraction*

The features extracted by the second channel can be expressed as

$$F_l = f_l(X; \theta_l)$$

where $f_l$ is the function representing the operations in the second channel (e.g., BDR), $\theta_l$ represents the parameters of the second channel, and $F_l$ is the extracted feature set from $X$.

### 2. Feature Aggregation
The features from both channels are combined into a single feature vector $F$:

$$F = \text{concat}(F_h, F_l)$$

This step involves concatenating (or other methods of aggregation like summing) the feature vectors $F_h$ and $F_l$.

### 3. Dense Layer
The aggregated feature vector $F$ is then processed through a dense layer to create an intermediate vector $Z$:

$$Z = W \cdot F + b$$

where $W$ is the weight matrix of the dense layer, $b$ is the bias vector, and $Z$ might be processed by an activation function (e.g., ReLU).

### 4. Classification or Regression Layer
Depending on the specific task (classification or regression), the final output is computed from $Z$. For classification tasks, typically with $k$ classes:

$$P = \text{Softmax}(W_c \cdot Z + b_c)$$

where $W_c$ and $b_c$ are the weights and biases of the classification layer, and $P$ represents the probability distribution over the classes.

For regression tasks:

$$Y = W_r \cdot Z + b_r$$

where $W_r$ and $b_r$ are the weights and biases of the regression layer, and $Y$ is the predicted output values.

### Extraction of high- and low-dimensional features

The HD features are extracted using traditional deep learning layers such as convolution, max-pooling, batch normalization, and fully connected layers. The LD features are extracted using BDR approach as described below.

### Bayesian dimensionality reduction

To preserve the manifold structure of input data/features in LD representation, we use an unsupervised Bayesian projection method[13]. The technique is based on the following joint data distribution

$$p(\boldsymbol{\Theta}, \boldsymbol{\Xi} | \boldsymbol{X}) = p(\boldsymbol{\Phi}) \, p(\boldsymbol{Q} | \boldsymbol{\Phi}) \, p(\boldsymbol{Z} | \boldsymbol{Q}, \boldsymbol{X}), \tag{1}$$

where $\boldsymbol{X}$ (of size $D \times N$) denotes the input data, $\boldsymbol{Q}$ (of size $D \times R$) is the projection matrix, and $\boldsymbol{Z}$ (of size $R \times N$) are the latent variables. The prior variables are denoted by $\boldsymbol{\Xi} = \{\boldsymbol{\Phi}\}$, and the remaining variables are denoted by $\boldsymbol{\Theta} = \{\boldsymbol{Q}, \boldsymbol{Z}\}$. The hyperparameters are denoted by $\zeta = \{\alpha_\phi, \beta_\phi, \sigma_z^2\}$. See supplementary Tables S1 and S2 for a list of the notations and probability distribution functions for each of the variables.

The posterior distribution can be approximated as follows

$$p(\boldsymbol{\Theta}, \boldsymbol{\Xi} | \boldsymbol{X}) \approx q(\boldsymbol{\Theta}, \boldsymbol{\Xi}) = q(\boldsymbol{\Phi}) \, q(\boldsymbol{Q}) \, q(\boldsymbol{Z}), \tag{2}$$

where the factored posterior can be modeled as a product of gamma distributions for the precision parameters and Gaussian distributions for the projection matrix and latent variables.

#### Case 1: Element-wise Prior

For the element-wise prior over $\boldsymbol{\Phi}$, the approximate posterior of the precision parameters is

$$q(\boldsymbol{\Phi}) = \prod_{f=1}^{D} \prod_{s=1}^{R} \mathscr{G}\left(\phi_s^f; \alpha_\phi + \frac{1}{2}, \left(\frac{1}{\beta_\phi} + \frac{\left(q_s^f\right)^2}{2} + \frac{1}{2}\sigma_{q_s}^{ff}\right)^{-1}\right), \tag{3}$$

where $\mathscr{G}(\cdot; \alpha, \beta)$ denotes the gamma distribution with shape parameter $\alpha$ and scale parameter $\beta$, $q_s^f$ denotes the $f$-th element of the $s$-th column of $\boldsymbol{Q}$, and $\sigma_{q_s}^{ff}$ is the $f$-th diagonal element of the covariance matrix $\Sigma_{\boldsymbol{q}_s}$.

The approximate posterior distribution of the projection matrix $\boldsymbol{Q}$ is a product of multivariate Gaussian distributions

$$q(\boldsymbol{Q}) = \prod_{s=1}^{R} \mathscr{N}\left(\boldsymbol{q}_s; \Sigma_{\boldsymbol{q}_s} \boldsymbol{X} \widetilde{\boldsymbol{z}}^s / \sigma_z^2, \Sigma_{\boldsymbol{q}_s}\right), \tag{4}$$

where:

$$\Sigma_{\boldsymbol{q}_s} = \left(\operatorname{diag}\left(\widetilde{\boldsymbol{\phi}}_s\right) + \frac{\boldsymbol{X}\boldsymbol{X}^\top}{\sigma_z^2}\right)^{-1}, \tag{5}$$

and $\widetilde{\boldsymbol{\phi}}_s$ denotes the posterior expectation of $\boldsymbol{\phi}_s$, $\widetilde{\boldsymbol{z}}^s$ denotes the $s$-th row vector of the mean matrix of $\boldsymbol{Z}$, and $\operatorname{diag}(\cdot)$ creates a diagonal matrix from a vector.

#### Case 2: ARD Prior

For the Automatic Relevance Determination (ARD) prior, the precision parameters are shared across dimensions

$$q(\boldsymbol{\phi}) = \prod_{s=1}^{R} \mathscr{G}\left(\phi_s; \alpha_\phi + \frac{D}{2}, \left(\frac{1}{\beta_\phi} + \frac{1}{2}\left(\boldsymbol{q}_s^\top \boldsymbol{q}_s + \operatorname{Tr}(\Sigma_{\boldsymbol{q}_s})\right)\right)^{-1}\right), \tag{6}$$

where $\operatorname{Tr}(\cdot)$ denotes the trace of a matrix.

The approximate posterior distribution of the projection matrix remains as in Equation (4), but with the covariance matrix

$$\Sigma_{\boldsymbol{q}_s} = \left(\widetilde{\phi}_s \boldsymbol{I}_D + \frac{\boldsymbol{X}\boldsymbol{X}^\top}{\sigma_z^2}\right)^{-1}, \tag{7}$$

where $\widetilde{\phi}_s$ is the posterior expectation of $\phi_s$ and $\boldsymbol{I}_D$ is identity matrix of size $D \times D$. The term $\boldsymbol{X}\boldsymbol{X}^\top$ is included in the covariance of $\boldsymbol{Q}$ to ensure that the variance of the original data and that of the projected matrix remain similar. This is analogous to Principal Component Analysis (PCA), which also attempts to preserve the maximum variance of data in principal directions.

**Approximate Posterior of Latent Variables**

The approximate posterior distribution of the latent variables $\boldsymbol{Z}$ is a multivariate Gaussian distribution

$$q(\boldsymbol{Z}) = \prod_{n=1}^{N} \mathcal{N}\left(\boldsymbol{z}_n; \Sigma_{\boldsymbol{z}} \boldsymbol{Q}^\top \boldsymbol{x}_n / \sigma_z^2, \Sigma_{\boldsymbol{z}}\right), \tag{8}$$

where

$$\Sigma_{\boldsymbol{z}} = \left(\frac{\boldsymbol{Q}^\top \boldsymbol{Q}}{\sigma_z^2} + \mathbf{I}_R\right)^{-1}, \tag{9}$$

and $\boldsymbol{z}_n$ denotes the $n$-th column vector of $\boldsymbol{Z}$, $\boldsymbol{x}_n$ denotes the $n$-th column vector of $\boldsymbol{X}$, and $\mathbf{I}_R$ is an identity matrix of size $R \times R$.

**QR Decomposition of $Q_\mu$**

Given the matrix $Q_\mu = \mathbb{E}_{q(\boldsymbol{Q})}[\boldsymbol{Q}] \in \mathbb{R}^{D \times R}$, we perform the QR decomposition to obtain

$$Q_\mu = \mathbf{Q}_{\text{mu\_orth}} \mathbf{R}, \tag{10}$$

where $\mathbf{Q}_{\text{mu\_orth}} \in \mathbb{R}^{D \times R}$ is a matrix with orthonormal columns satisfying $\mathbf{Q}_{\text{mu\_orth}}^\top \mathbf{Q}_{\text{mu\_orth}} = \mathbf{I}_R$ and $\mathbf{R} \in \mathbb{R}^{R \times R}$ is an upper triangular matrix.

**Projection Computation**

The projection of the data is computed by

$$\boldsymbol{U} = \boldsymbol{X}^\top \mathbf{Q}_{\text{mu\_orth}}. \tag{11}$$

**Implementation and parameter settings**

Both Python and Matlab 2024a versions of DWL implementation are available. The deep learning models for different tasks were implemented and trained in MATLAB and Python. PCA and t-SNE implemented by Matlab have been used to produce the results of these method with default parameters. All the scale parameters ($\alpha_\lambda, \beta_\lambda, \alpha_\phi, \beta_\phi, \alpha_\psi$, and $\beta_\psi$) in Bayesian projection in BDR were initialized as 1. The number of maximum iterations was set to 200. Matlab's default settings were used for specification of different layers of the analyzed DNNs. In the training process of DNNs, mini-batch size was set to 32 in all trainings and validation frequency was set to the closest integer to the total number of training images divided by the mini-batch size. Maximum epoch was set to 1000 and shuffling was performed in every epoch in all trainings. ADAM optimizer was used for network optimization. Initial learning rate was fixed at 0.002[37] and cross entropy loss was used in all training. All the networks were initialized using Kaiming initialization method[38].

Following the studies in Refs.[39,40] and common practices in medical image analysis community, we randomly divided the datasets into training, validation and testing in 70%:15%:15% ratio. For image datasets, data augmentation is applied on the training dataset with random reflection, translation and scaling along the X and Y directions. For translation operation, the pixel range was set to $-30$ to $30$ and for scaling operation, scale range was set from 0.9 to 1.1. We used early stopping by the validation performance for training all the DNNs. The validation accuracy and loss are checked at each epoch, and the training is terminated if the validation accuracy does not improve for 10 consecutive epochs (patience parameter of 10 in Keras EarlyStopping function[41])[42,43]. The models with the best validation accuracy are saved and used in our analyses.

**Statistics & Reproducibility**

No statistical method was used to predetermine sample size. No data were excluded from the analyses. The experiments were not randomized. There was no blinding. The analyses performed do not involve evaluation of any subjective matters.

# Acknowledgments

This work was partially supported by NIH (1R01 CA256890, R01CA275772, and 1K99LM014309), and an Innovation Award from Stanford Cancer Institute.

# Author contributions statement

L.X. and M.T.I conceived the experiment(s), M.T.I conducted the experiment(s). Both authors reviewed the manuscript.

## Competing interests

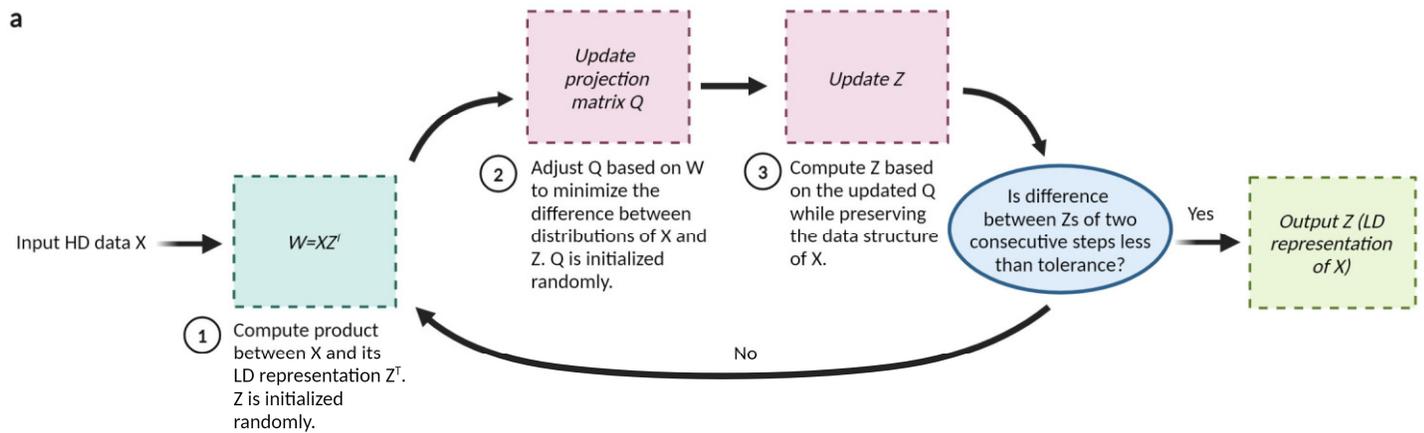

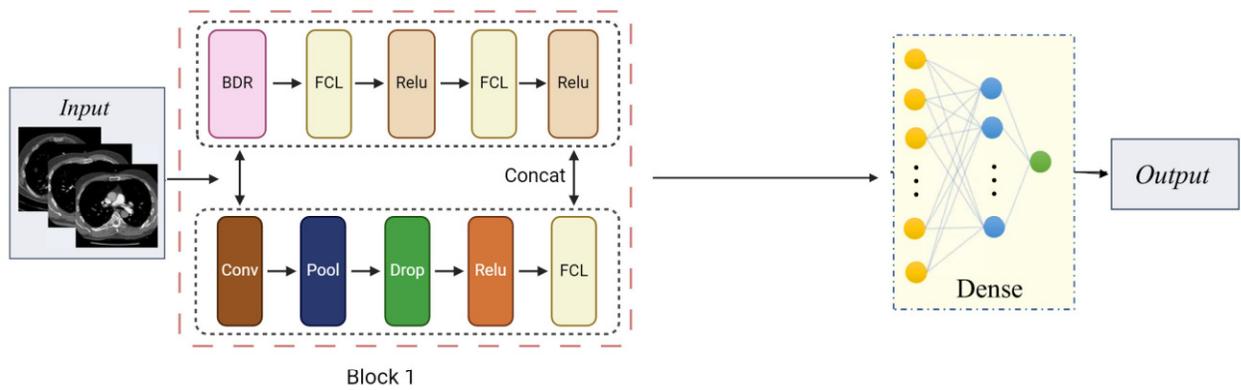

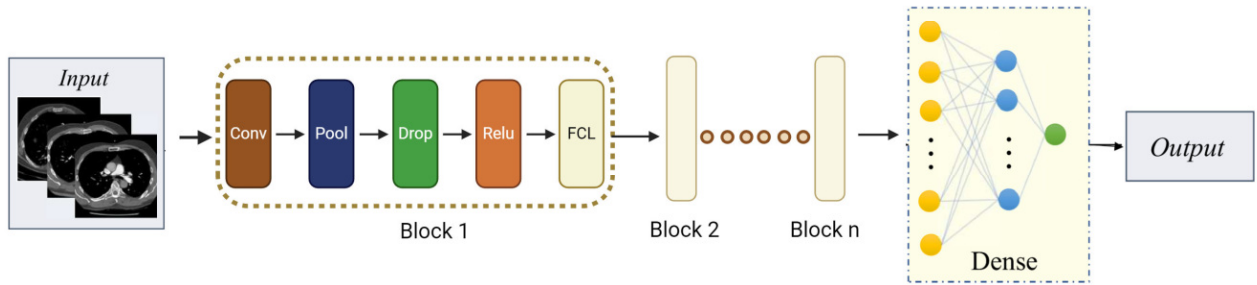

**Figure 1.** (a) Workflow of Bayesian dimensionality reduction (BDR) (b) Architecture of dual-interactive-channel network (D-Net). D-Net has two components that work in harmony to enhance performance and interpretability. The first consists of convolutional layers, which are responsible for extracting HD spatial features (intra-data relationships) from the input data. The second one uses BDR to extract the LD feature representation by leveraging inter-data relationships and facilitates the information flow.

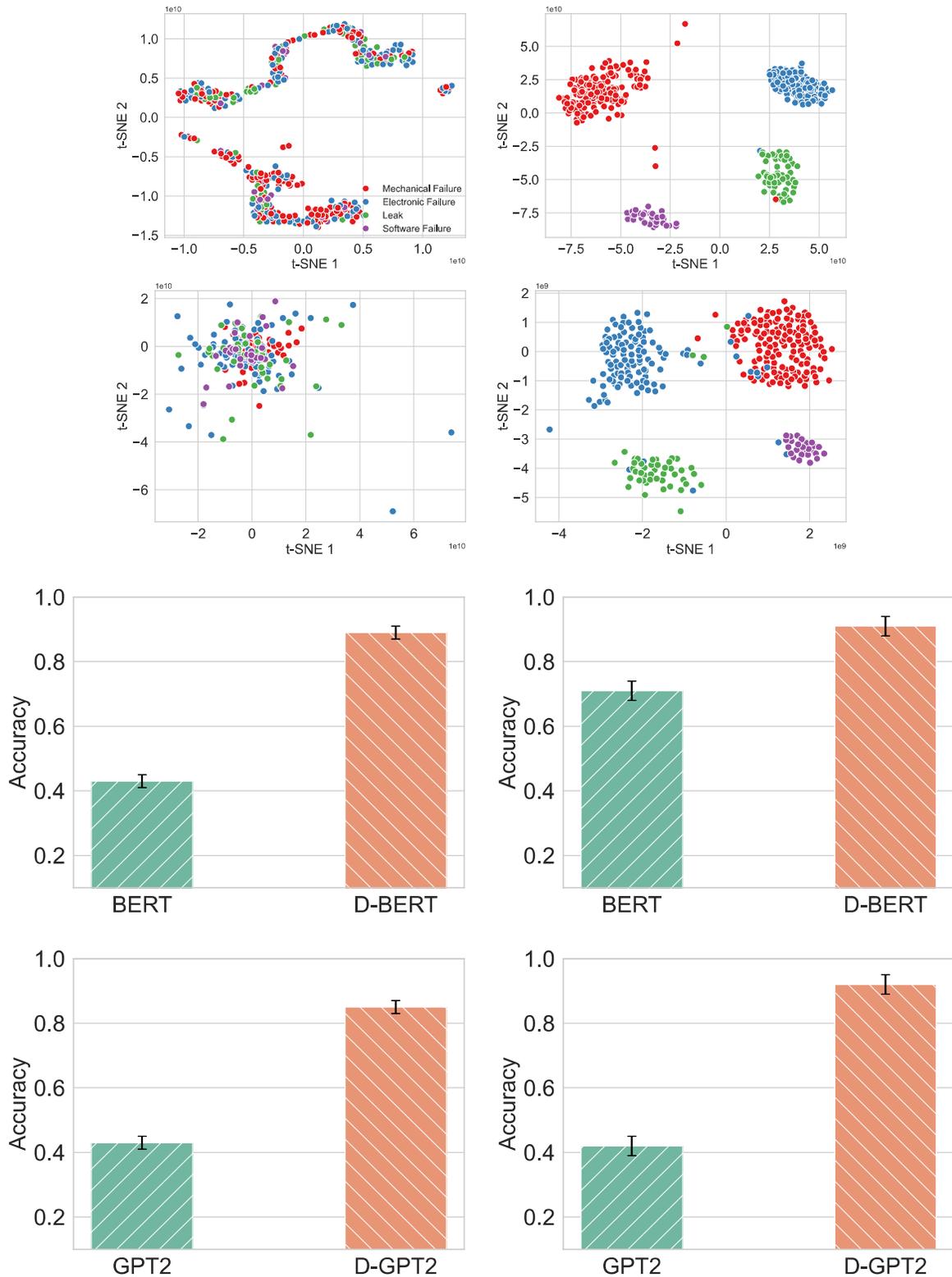

**Figure 2.** Text classification using BERT and GPT2 and their D-Net counterparts. (Row-1) t-SNE visualizations of the last fully connected layers of BERT and D-BERT classifiers. (Row-2) t-SNE visualizations of the last fully connected layers of GPT2 and D-GPT2 classifiers. From both these visualizations, it is seen that features are much better clustered into classes in cases of D-BERT and D-GPT2. (Row-3) Classification accuracy of the testing data using BERT and D-BERT after epochs 10 and 2,000 (from left to right). (Row-4) Classification accuracy of the testing data using GPT2 and D-GPT2 after epochs 10 and 2,000 (from left to right).

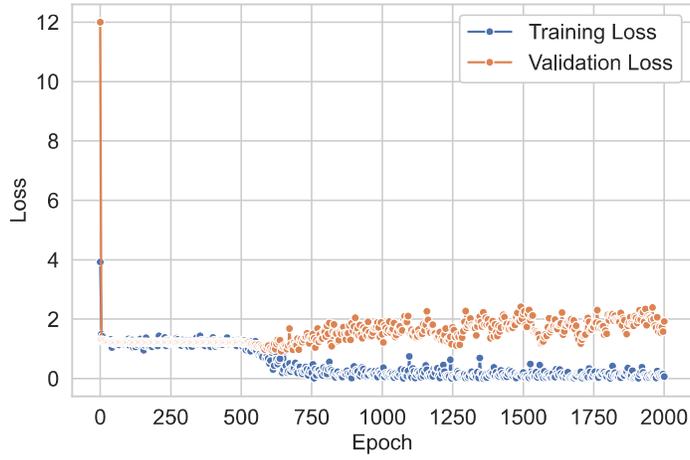
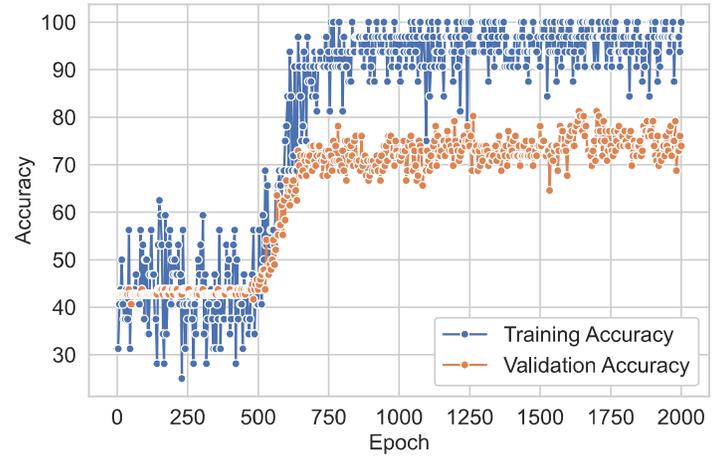
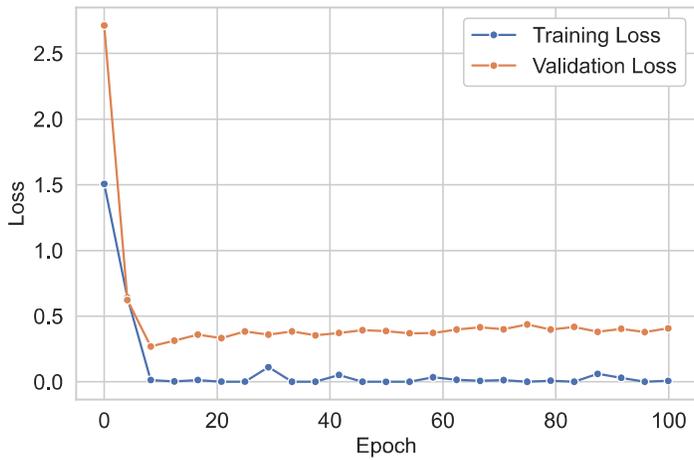
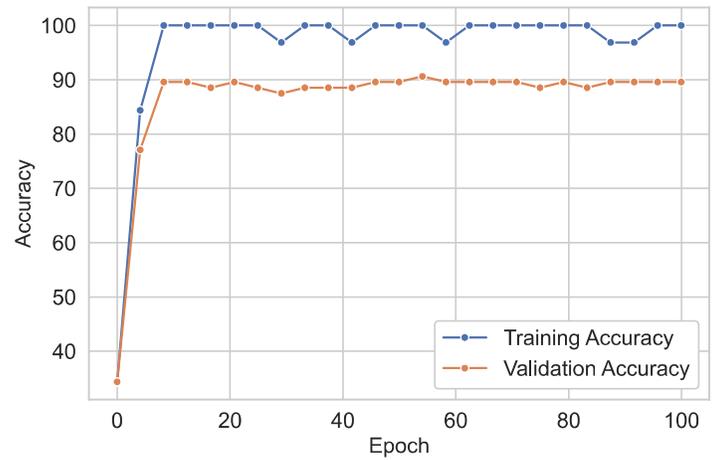

**Figure 3.** (Row-1) Training and validation loss and accuracy of BERT (row-2) Training and validation loss and accuracy of D-BERT.

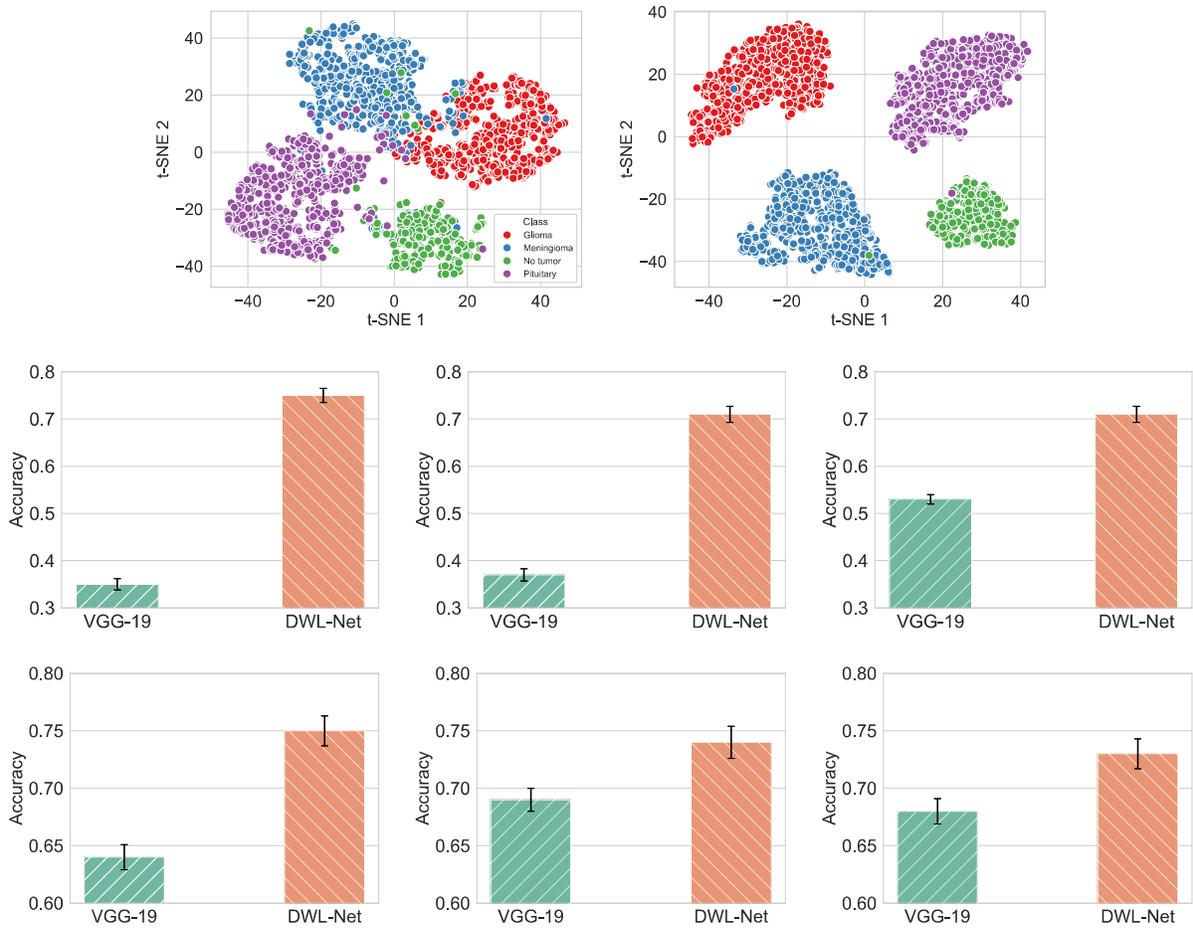

**Figure 4.** Brain tumor classification from MRI images. (Row-1) t-SNE visualizations of the last fully connected layers of VGG-19 and D-Net. From these visualizations, it is seen that features are much better clustered into classes in D-Net. (Row-2) Classification accuracy of the testing data using VGG-19 and D-Net after epoch 5 for three different layer numbers (30, 35 and 40 from left to right). (Row-3) Classification accuracy of the testing data using VGG-19 and D-Net after epoch 50 for three different layer numbers (30, 35 and 40 from left to right).

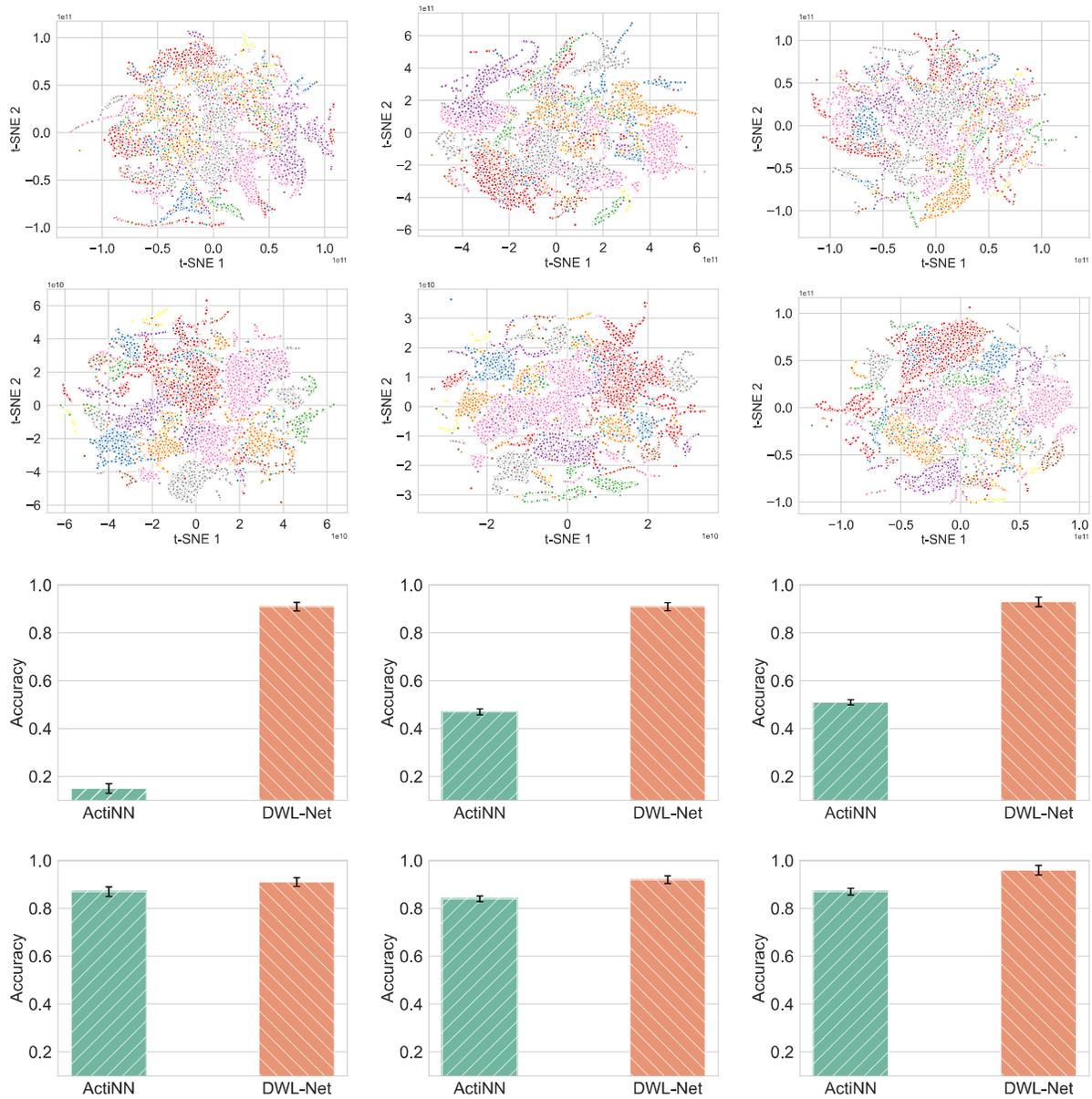

**Figure 5.** Cell annotation using state-of-the-art DNN (ActiNN) and D-Net. (Row-1) t-SNE visualizations of the features extracted from the training data from the last three FCLs of ActiNN. (Row 2) t-SNE visualizations of the features extracted from the training data from the last three FCLs of D-Net. From these visualizations, it is evident that the features are much better clustered in D-Net. (Row-3) Classification accuracy of the testing data using ActiNN and D-Net after epoch 5 for three percent of training data (50, 70 and 90 from left to right). (Row-4) Classification accuracy of the testing data using ActiNN and D-Net after epoch 50 for three percent of training data (50, 70 and 90 from left to right).

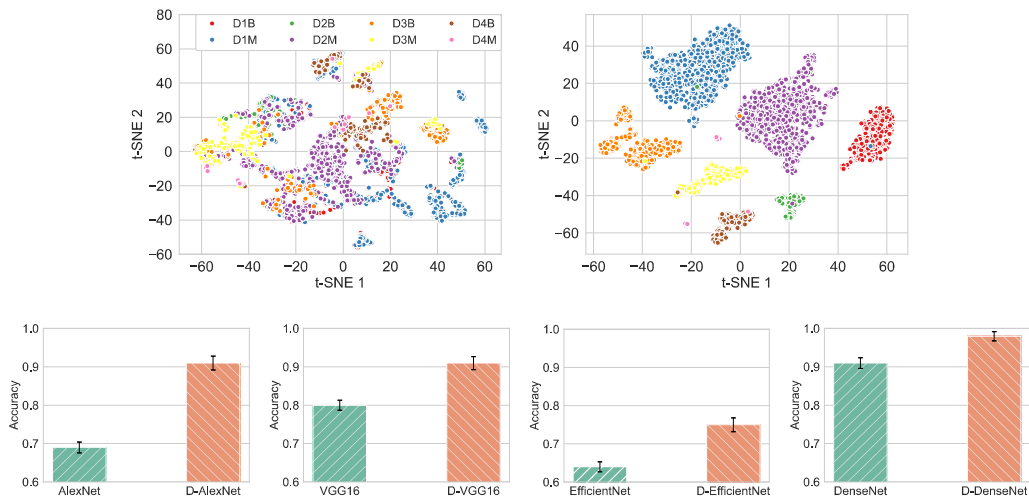

**Figure 6.** Breast tumor classification from mammography images. (Row-1) t-SNE visualizations of the final FCLs of VGG-16 and D-Net. From these visualizations, it is seen that features are much better clustered in D-Net. (Row-2) Classification accuracy of the testing data using original AlexNet, VGG-16, EfficientNet and DenseNet and their D-Net counterparts (from left to right).

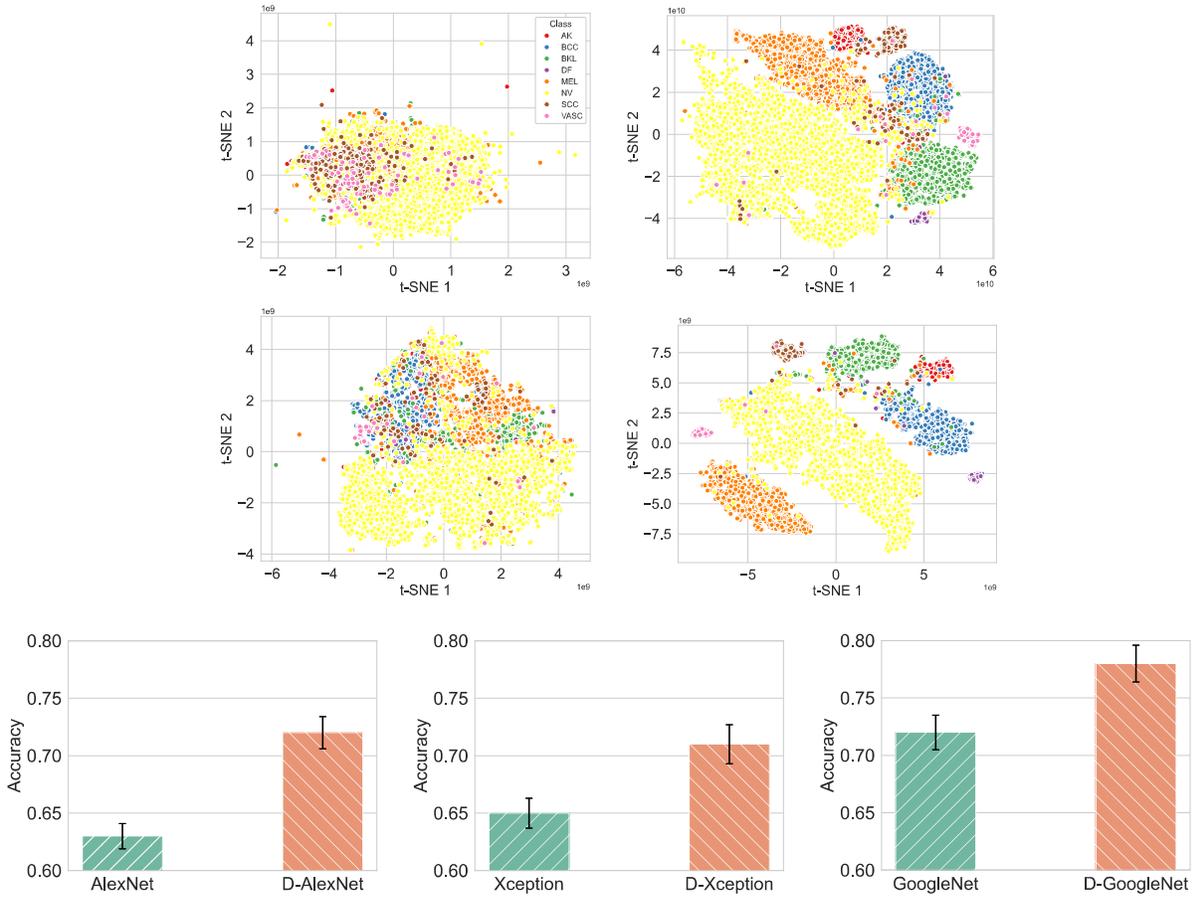

**Figure 7.** Skin cancer classification from ISIC images. (Row-1) t-SNE visualizations of the features extracted from training data from the final FCLs of AlexNet (column 1) and D-AlexNet (column 2). (Row-2) t-SNE visualizations of the features extracted from testing data from the final FCLs of AlexNet (column 1) and D-AlexNet (column 2). From these visualizations, it can be observed that the features are much better clustered in D-Net. (Row-3) Classification accuracy of the testing data using original AlexNet, Xception, and GoogleNet and their D-Net counterparts (from left to right).

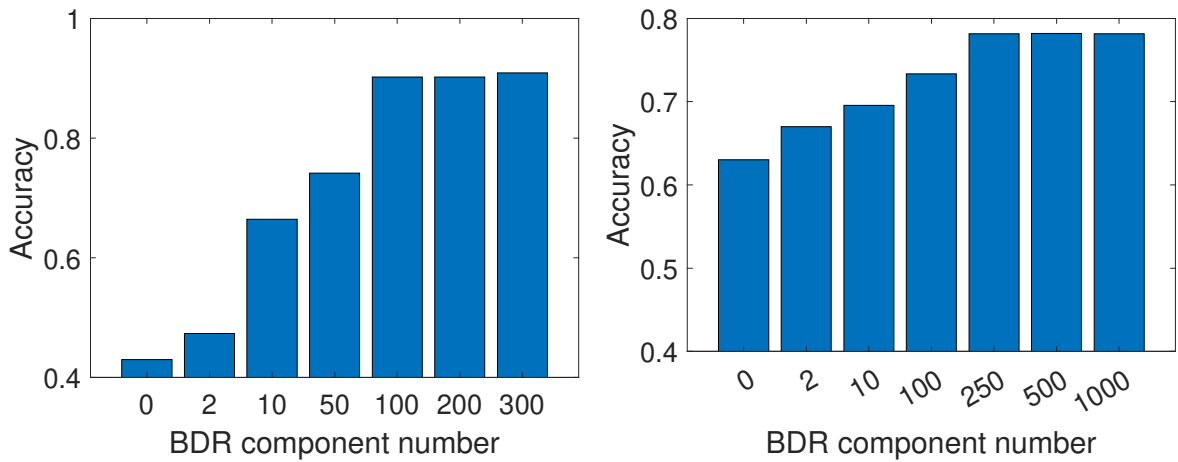

**Figure 8.** Influence of the number of LD components on DWL performance. Classification accuracy of D-GPT2 and D-VGG19 for (left) text classification task and (right) tumor classification task for different numbers of BDR component.